\documentclass[manuscript,screen,review，nonatbib]{acmart}

\renewcommand\footnotetextcopyrightpermission[1]{} 
\usepackage{multirow,amsmath,booktabs,diagbox}
\usepackage{algorithmic}
\usepackage{algorithm}

\newcommand{\Highlight}[1]{\textcolor{black}{#1}}
\newcommand{\highlight}[1]{\textcolor{black}{#1}}

\newcommand{\hhhighlight}[1]{\textcolor{black}{#1}}
\newcommand{\hlight}[1]{\textcolor{black}{#1}}
\newcommand{\etal}{\textit{et al.}}

\AtBeginDocument{%
  }

\setcopyright{acmlicensed}
\copyrightyear{2024}
\acmYear{2024}
\acmDOI{XXXXXXX.XXXXXXX}

\acmConference[Conference acronym 'XX]{Make sure to enter the correct
  conference title from your rights confirmation emai}{June 03--05,
  2018}{Woodstock, NY}
\acmISBN{978-1-4503-XXXX-X/18/06}

\begin{document}

\title{Micro-Expression Recognition via Fine-Grained Dynamic Perception}

\author{Zhiwen~Shao, Yifan~Cheng}
\affiliation{%
  \institution{School of Computer Science and Technology, China University of Mining and Technology, Xuzhou, China, and Mine Digitization Engineering Research Center of the Ministry of Education, Xuzhou, China, and also Department of Computer Science and Engineering, The Hong Kong University of Science and Technology, Hong Kong}
  \country{China}
}
\email{{zhiwen\_shao; yifan\_cheng}@cumt.edu.cn}





\author{Fan~Zhang}
\affiliation{%
 \institution{Inspur Zhuoshu Big Data Industry Development Co., Ltd., Jinan}
 \country{China}}
\email{zhangfan\_inspur@foxmail.com}

\author{Xuehuai~Shi}
\affiliation{%
  \institution{School of Computer Science, Nanjing University of Posts and Telecommunications, Nanjing}
  \country{China}}
\email{xuehuai@njupt.edu.cn}

\author{Canlin~Li}
\affiliation{%
 \institution{School of Computer Science and Technology, Zhengzhou University of Light Industry, Zhengzhou}
 \country{China}}
 \email{li-cl@zzuli.edu.cn}

\author{Lizhuang~Ma}
\affiliation{%
  \institution{School of Computer Science, Shanghai Jiao Tong University, Shanghai}
  \country{China}}
  \email{ma-lz@cs.sjtu.edu.cn}

\author{Dit-Yan~Yeung}
\authornote{Corresponding authors: Yifan~Cheng, Xuehuai~Shi, and Dit-Yan~Yeung.}
\affiliation{%
  \institution{Department of Computer Science and Engineering, The Hong Kong University of Science and Technology, Hong Kong}
  \country{China}}
\email{dyyeung@cse.ust.hk}

\renewcommand{\shortauthors}{Z. Shao et al.}

\begin{abstract}
Facial micro-expression recognition (MER) is a challenging task, due to the transience, subtlety, and dynamics of micro-expressions (MEs). Most existing methods resort to hand-crafted features or deep networks, in which the former often additionally requires key frames, and the latter suffers from small-scale and low-diversity training data.
In this paper, we develop a novel fine-grained dynamic perception (FDP) framework for MER. We propose to rank frame-level features of a sequence of raw frames in chronological order, in which the rank process encodes the dynamic information of both ME appearances and motions. Specifically, a novel local-global feature-aware transformer is proposed for frame representation learning. A rank scorer is further adopted to calculate rank scores of each frame-level feature. Afterwards, the rank features from rank scorer are pooled in temporal dimension to capture dynamic representation. Finally, the dynamic representation is shared by a MER module and a dynamic image construction module, in which the former predicts the ME category, and the latter uses an encoder-decoder structure to construct the dynamic image. The design of dynamic image construction task is beneficial for capturing facial subtle actions associated with MEs and alleviating the data scarcity issue. Extensive experiments show that our method (i) significantly outperforms the state-of-the-art MER methods, and (ii) works well for dynamic image construction. \hlight{Particularly, our FDP improves by 4.05\%, 2.50\%, 7.71\%, and 2.11\% over the previous best results in terms of F1-score on the CASME II, SAMM, CAS(ME)$^2$, and CAS(ME)$^3$ datasets, respectively.} The code is available at \url{https://github.com/CYF-cuber/FDP}.
\end{abstract}





\maketitle

\section{Introduction}\label{sec1}
Facial micro-expression recognition (MER) has recently gained increasing attention in the fields of computer vision and affective computing~\cite{xie2020assisted,kumar2021micro,shao2019facial,shao2021explicit}. \hhhighlight{It has} applications in many areas, \hhhighlight{as micro-expressions (MEs) can reveal emotions those are attempted to conceal~\cite{ekman2009lie}.} 
\hhhighlight{For instance, in mental health, MER can be used to spot signs of disorders like depression and monitors treatment progress. It can also be used to detect non-verbal pain in patients who cannot communicate well. Recently, Zhou \etal~\cite{zhou2024multi} proposed a multi-modal fine-grained depression detection method via fusing audio and text features. However, visual modality ME that can well reflect the degree of depression is ignored. In public security, MER can be used to aid criminal investigations in terms of lie detection and suspect identification. It can also be used to spot suspicious emotional states in the crowd so as to prevent threats. However, MEs are often neglected in recent lie detection works~\cite{javaid2022eeg,hossain2024mvis4ld}. Therefore, we explore a new MER solution to empower mental health and public security.}
MEs are facial subtle muscle actions, and are dynamic during a short duration with no more than 500 milliseconds~\cite{yan2013fast}. Besides, most of the existing ME datasets are small-scale~\cite{yan2014casme,davison2016samm}, due to the large costs of manual labeling. With limited training data to capture challenging MEs, MER remains a difficult task.

One common solution is to adopt hand-crafted features associated with MEs. These features typically try to capture motion patterns~\cite{zach2007duality,chaudhry2009histograms}, encode spatio-temporal information~\cite{zhao2007dynamic,bilen2017action}, or focus on local contrast information~\cite{davison_micro-facial_2015,li2018towards}. However, hand-crafted features based on prior knowledge only process partial characteristics, and have limited capacity to model challenging MEs in diverse samples. Moreover, these features like optical flow~\cite{zach2007duality} often additionally rely on key frames including onset, apex, and offset frames of MEs to improve the recognition performance, which limits the applicability.

\begin{figure}
\centering\includegraphics[width=0.88\linewidth]{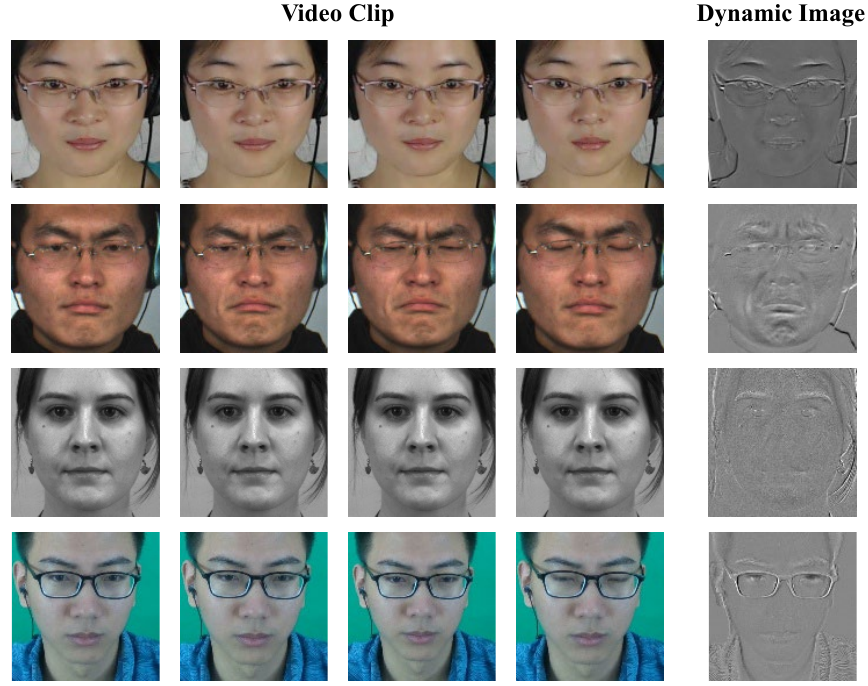}
\caption{Illustration of dynamic images~\cite{bilen2017action} for several example video clips. Each row shows four sample frames of a ME video clip as well as the generated dynamic image. \textit{The overall facial appearances and the highlighted motion areas can be observed from the dynamic images.}
}
\label{fig:dynamic_example}
\end{figure}

Another alternative way is to use prevailing deep neural networks. Zhou \etal~\cite{zhou2019dual} computed the optical flow between onset and apex frames of the input video, and then fed horizontal and vertical components of the optical flow into a dual-inception network to predict the ME category. However, pre-extracted optical flow as well as key frames are required. Some other methods directly input raw frame images to deep networks so as to remove the limitations of hand-crafted features. For example, Reddy \etal~\cite{reddy2019spontaneous} employed a 3D convolutional neural network (CNN) to capture spatial and temporal information, and Xia \etal~\cite{xia2021micro} used macro-expression recognition to facilitate MER. However, the capture of fine-grained ME information is not explicitly handled, and these methods suffer from insufficient training data.

To tackle the above issues, we introduce a \Highlight{rank pooling technique~\cite{fernando2016rank} to perceive temporal evolution of appearance in ME videos, and improve the transformer~\cite{vaswani_attention_2017} structure to capture both local and global characteristics.} 
\hhhighlight{As mentioned in~\cite{YAN2024130}, dynamic facial expression recognition has wider practical applications, such as empowering smart cities. Our work is based on dynamic and continuous ME frame sequence rather than pre-annotated key frames.}
\Highlight{By the rank pooling,} the spatial appearances and temporal motions of a video can be encoded as a dynamic image~\cite{bilen2017action}, which indicates the correlations between dynamic image and ME, as illustrated in Fig.~\ref{fig:dynamic_example}. In particular, we propose an end-to-end \hhhighlight{F}ine-grained \hhhighlight{D}ynamic \hhhighlight{P}erception framework called \textbf{FDP}, which jointly estimates ME and \Highlight{constructs} dynamic image of the input video. First, a novel local-global feature-aware transformer is proposed to capture ME related local information while preserving the global relational modeling ability of vanilla transformer~\cite{vaswani_attention_2017} \highlight{for single frame representation learning}.
\Highlight{After extracting the local-global feature, a rank scorer is further employed to learning temporal rank information of each frame. 
Then, a 3D CNN based temporal pooling module is applied to capture temporal features from all the single rank features so as to learn video-wide dynamic representation.}
Finally, two modules of MER estimation and dynamic image construction are adopted to predict the ME category and the dynamic image, respectively. 

The main contributions of this work are threefold:
    
    $\bullet$ We propose a novel fine-grained dynamic perception framework with MER and dynamic image construction, which does not depend on pre-extracted hand-crafted features and key frames. The use of dynamic image construction task contributes to capturing facial subtle muscle actions related to MEs, which relaxes the dependence of our deep network on large-scale training samples.
    
    
     $\bullet$ We propose a novel local-global feature-aware transformer to capture local subtle information associated with MEs while preserving the global modeling capacity of transformer.
    
     $\bullet$ Extensive experiments on CASME II, SAMM, {CAS(ME)$^2$, and CAS(ME)$^3$}  benchmarks demonstrate that our approach soundly outperforms the state-of-the-art MER methods, and achieves competitive performance for dynamic image construction.

\section{Related Work}\label{sec2}

In this section, we review the previous methods those are closely associated with our approach, including non-deep learning based MER, deep learning based MER, \Highlight{rank pooling and dynamic image, }and combination of CNN and vision transformer (ViT).

\subsection{Non-Deep Learning Based MER}

Since MEs are subtle and hardly distinguishable, earlier methods propose hand-crafted features based on prior knowledge about local characteristics and motion patterns. Zhao \etal~\cite{zhao2007dynamic} designed local binary patterns from three orthogonal planes (LBP-TOP) by considering co-occurrence statistics of motions in three directions. Wang \etal~\cite{wang2015efficient} further proposed local binary patterns with six intersection points (LBP-SIP) to avoid duplicated encoding in LBP-TOP. Ben \etal~\cite{ben2018learning} proposed binary face descriptors including dual-cross patterns from three orthogonal planes (DCP-TOP) and hot wheel patterns from three orthogonal planes (HWP-TOP) to encode the discriminative features of ME videos. Another solution of hand-crafted features is based on histogram. Davison \etal~\cite{davison_micro-facial_2015} designed histogram of oriented gradients (HOG), and Li \etal~\cite{li2018towards} further proposed histogram of image gradient orientation (HIGO).

Besides, optical flow describes the motion pattern of each pixel across frames, which has been widely used in MER. Davison \etal~\cite{liong2018less} proposed bi-weighted oriented optical flow (Bi-WOOF) by using onset frame and apex frame to represent a ME. Happy \etal~\cite{happy2019fuzzy} developed histogram of oriented optical flow (HOOF)~\cite{chaudhry2009histograms} to FHOOF by using fuzzy membership function to collect motion directions, and further developed FHOOF to fuzzy histogram of optical flow orientations (FHOFO) by ignoring subtle motion magnitudes. Dynamic image~\cite{bilen2017action} is another newer way to encode facial motion information of a video, which has been introduced to MER~\cite{verma2020learnet,nie2021geme}.

However, these hand-crafted features only focus on partial characteristics associated with MEs, and often additionally rely on key frames of MEs.

\subsection{Deep Learning Based MER}

Considering the power of deep neural networks~\cite{shao2021jaa,shao2023facial,shao2024facial}, Reddy \etal~\cite{reddy2019spontaneous} introduced a 3D CNN to capture spatial and temporal information from raw image sequences for MER. Wei \etal~\cite{wei2022novel} proposed an attention-based magnification-adaptive network (AMAN) to magnify and focus on ME details. Since subtle MEs are hard to capture, some methods adopt correlated tasks to facilitate MER. Xie \etal~\cite{xie2020assisted} proposed an AU-assisted graph attention convolutional network (AU-GACN) to reason the relationships among AUs so as to assist the recognition of MEs. Xia \etal~\cite{xia2021micro} introduced macro-expression recognition as an auxiliary task, and used adversarial learning to align the feature distributions between macro-expressions and MEs. 

Since current deep networks suffer from small-scale and low-diversity ME datasets, other approaches combine hand-crafted features with deep learning. Hu \etal~\cite{hu2018multi} incorporated local Gabor binary pattern from three orthogonal panels (LGBP-TOP) features and CNN features, and then trained MER by treating the classification of each ME category as a one-against-all classification problem. 
\hhhighlight{Liu \etal~\cite{liu2022micro} extracted TV-L1 optical flow between key frames to input into a pre-trained ShuffleNet and then conducted classification via support vector machine (SVM)}.
Verma \etal~\cite{verma2020learnet} first extracted the dynamic image of the input ME video, and then fed it into a lateral accretive hybrid network (LEARNet). Shao \etal~\cite{shao2023identity} generated the optical flow between onset and apex frames of the input video, then input horizontal and vertical optical flow components to a dual-inception network, and finally jointly train MER and AU recognition based on a transformer~\cite{vaswani_attention_2017}.

All these methods suffer from insufficient training data, or dependence on hand-crafted features. In contrast, our method put MER and dynamic image construction into a joint learning framework, in which raw images are handled, and the auxiliary task alleviate the requirement of large-scale training data.

\subsection{\Highlight{Rank Pooling and Dynamic Image}}
\Highlight{Rank pooling~\cite{fernando2016rank} is a video representation technique used in video analysis to aggregate information over time, typically in action recognition tasks~\cite{cherian2017generalized, bilen2017action}. It aims to summarize a sequence of ranked frames into a single representative feature vector. The process involves ranking frames, temporal pooling and optimization.}
\Highlight{Compared to rank pooling, dynamic image~\cite{bilen2017action} summarizes a whole video into a single image, which synthesizes a static representation that captures the motion dynamics. In the previous work~\cite{verma2020learnet}, dynamic image is employed as a pre-extracted feature to directly input into deep networks.}

\Highlight{Inspired by the above works, we design FDP using the same process as rank pooling, and further construct the dynamic image using rank features. In our approach, subtle ME actions are handled by learning video-wide temporal evolution, which includes ranking the frames in temporal dimension and constructing the dynamic image.}

\subsection{Combination of CNN and ViT}

In the past few years, CNN and ViT have achieved great performance successively in many vision tasks~\cite{lin2013network, he2016deep, arnab2021vivit, bertasius2021space}, in which the former works well in modeling local relationships and the latter is good at extracting global features. However, pure convolution struggles to capture long-range dependencies due to the limited reception field, and vanilla ViT that relies on self-attention mechanism is inefficient to encode low-level features.

Recently, hybrid structures of CNNs and ViTs are designed to improve the representation ability, in which local and global information are simultaneously focused while their respective weaknesses are avoided. 
Liu \etal~\cite{liu2020convtransformer} proposed a ConvTransformer with multi-head convolutional self-attention layers, to achieve video frame sequence learning and video frame synthesis.
Yuan \etal~\cite{yuan2021incorporating} combined the advantages of CNN and transformer, in which the former works well in extracting low-level features and strengthening locality, and the latter can establish long-range dependencies by extracting patches from low-level features and can promote the correlations among neighboring tokens in the spatial dimension.
Srinivas \etal~\cite{srinivas2021bottleneck} replaced the vanilla convolution with multi-head self-attention in the last several blocks of ResNet~\cite{he2016deep}.
Guo \etal~\cite{guo2022cmt} proposed a CMT network by
inheriting the merits of CNN and ViT, which is composed of depthwise convolutions with local perception units and a light-weight transformer module.

In our work, we integrate the merits of CNNs and ViTs by designing a local-global feature-aware transformer with local relational aggregator and global relational aggregator. Due to the capture of long-range
dependencies and local information, our method is effective at modeling transient, subtle, and dynamic MEs.

\begin{figure}
\centering\includegraphics[width=0.98\linewidth]{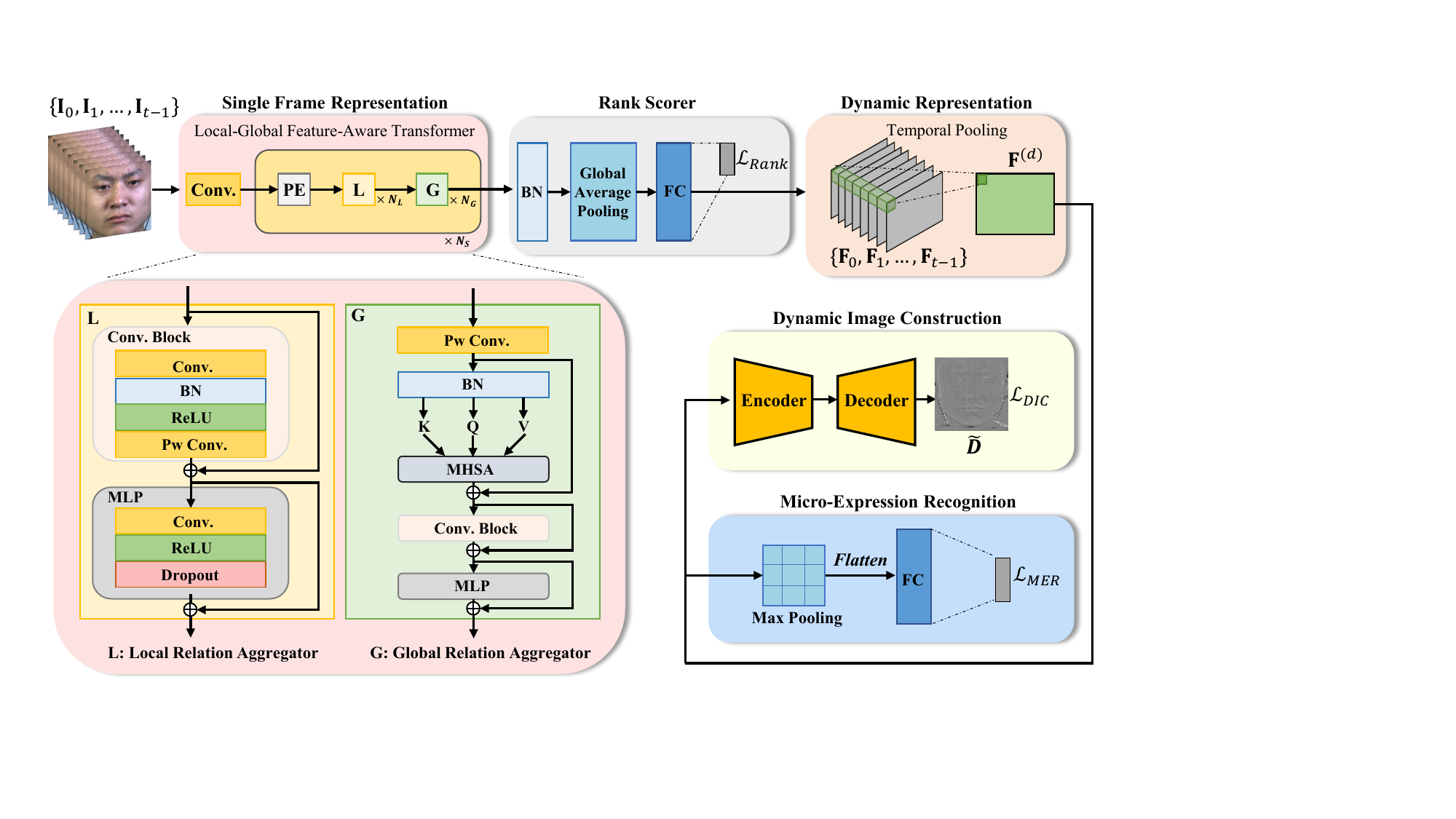} 
\caption{The architecture of our FDP. Given a sequence of $t$ frames $\{\mathbf{I}_{0}, \mathbf{I}_{1}, \cdots, \mathbf{I}_{t-1}\}$, FDP first extracts local-global feature $\mathbf{F}_k$ of each frame $\mathbf{I}_{k}$ by our proposed local-global feature-aware transformer.
\Highlight{Then, the local-global features $\{\mathbf{F}_{0}, \mathbf{F}_{1}, \cdots, \mathbf{F}_{t-1}\}$ are input to a fully connection layer based rank scorer to obtain the rank score of each frame, respectively.}
Afterwards, the sequence of local-global features $\{\mathbf{F}_{0}, \mathbf{F}_{1}, \cdots, \mathbf{F}_{t-1}\}$ is fed into a 3D convolutional layer to extract video-wide dynamic representation $\mathbf{F}^{(d)}$. Finally, $\mathbf{F}^{(d)}$ is fed into MER module and dynamic image construction module to estimate the ME category and the dynamic image $\widehat{\mathbf{D}}$, respectively. 
}
\label{fig:overview}
\end{figure}

\begin{algorithm}
\caption{\hhhighlight{The detailed processes of our FDP framework.}}
\label{alg:FDP}
\renewcommand{\algorithmicrequire}{\textbf{Input:}}
\renewcommand{\algorithmicensure}{\textbf{Output:}}
\begin{algorithmic}[1]
\REQUIRE A video clip $\{\mathbf{I}_{0}, \mathbf{I}_{1}, \cdots, \mathbf{I}_{t-1}\}$.
\ENSURE The predicted ME category $\hat{c}$ and dynamic image $\widehat{\mathbf{D}}$.
\STATE \textbf{Define} single frame representation module as $\mathcal{F}$. 
\STATE \textbf{Define} rank scorer as $\mathcal{R}$. 
\STATE \textbf{Define} dynamic representation module as $\mathcal{D}$. 
\STATE \textbf{Define} dynamic image construction module as $\mathcal{C}$.
\STATE \textbf{Define} micro-expression recognition module as $\mathcal{M}$. 
\FOR{each $k \in \{0,1, \cdots, t-1\}$}
 \STATE  Single frame representation $\mathbf{F}_{k} = \mathcal{F} (\mathbf{I}_{k} )$. 
\ENDFOR
 \STATE Rank loss $\mathcal{L}_{Rank}$ =   $\mathcal{R}$ ($\{\mathbf{F}_{0}, \mathbf{F}_{1}, \cdots, \mathbf{F}_{t-1}\}$), only used for training.
\STATE Dynamic representation $\mathbf{F}^{(d)}$ = $\mathcal{D}$ ($\{\mathbf{F}_{0}, \mathbf{F}_{1}, \cdots, \mathbf{F}_{t-1}\}$).
\STATE Predict ME category probabilities $\{\hat{p}_0,\cdots,\hat{p}_{m-1}\}$ =  $\mathcal{M}$ ($\mathbf{F}^{(d)}$), and obtain $\hat{c}=\mathop{\arg\max}\limits_{j \in \{0,1, \cdots, m-1\}} \hat{p}_j$.
\STATE $\widehat{\mathbf{D}}$ =  $\mathcal{C}$ ($\mathbf{F}^{(d)}$).
\STATE \textbf{Return} $\hat{c}$ and $\widehat{\mathbf{D}}$.
\end{algorithmic}
\end{algorithm}

\section{Rank Pooling Inspired Micro-Expression Recognition and Dynamic Image Construction}\label{sec3}

\subsection{\Highlight{Fine-Grained Dynamic Perception Framework}}

Given a video clip with $t$ frames $\{\mathbf{I}_{0}, \mathbf{I}_{1}, \cdots, \mathbf{I}_{t-1}\}$, we first obtain single frame representation $\mathbf{F}_k$ of the $k$-th frame $\mathbf{I}_{k}$ in the input video, respectively.
Then, a rank scorer $\mathbf{u}$ is employed to rate each single frame representation, in which the later frames are expected to receive higher scores according to the temporal order. Afterwards, the sequence of single frame representation $\{\mathbf{F}_{0}, \mathbf{F}_{1}, \cdots, \mathbf{F}_{t-1}\}$ is fed into a temporal pooling module to learn video-wide dynamic representation $\mathbf{F}^{(d)}$. Finally, $\mathbf{F}^{(d)}$ is shared by MER module and dynamic image construction module for joint learning.
Fig.~\ref{fig:overview} shows the overview of our framework, \hhhighlight{and Algorithm~\ref{alg:FDP} shows the detailed processes.}


Our FDP directly processes raw frame images without requiring key frames, and MER and dynamic image construction can contribute to each other in our joint learning framework. 

\subsection{Local-Global Feature-Aware Transformer}
\hhhighlight{Inspired by~\cite{FEI2025129323} that conducting local and global features simultaneously for facial expression recognition, we design a local-global feature extractor.} \Highlight{To learn high-quality frame representations while reducing the effect of noise and violent abrupt variations, we introduce local-global feature-aware transformer.} 
The goal of our proposed local-global feature-aware transformer is to capture local correlated information while modeling global dependencies. It consists of a vanilla convolutional layer, a stack of $N_S$ local-global relational aggregators, and a head block. Each local-global relational aggregator is a hybrid structure of CNN and ViT, which contains a patch embedding (PE) layer~\cite{dosovitskiy2021image}, a stack of $N_L$ CNN based local relational aggregators, and a stack of $N_G$ ViT based global relational aggregators. The head block is composed of a batch normalization (BN) layer~\cite{ioffe2015batch}and a global average pooling layer~\cite{lin2013network} to extract the final local-global feature. 

To enable the input of the first local-global relational aggregator, the patch embedding is obtained by extracting patches from the feature map of the vanilla convolutional layer. The subsequent patch embeddings are obtained from the output of the previous local-global relational aggregator.
We will elaborate the local relational aggregator and the global relational aggregator in the following sections.

\subsubsection{Local Relational Aggregator}

We design a local relational aggregator based on CNN while incorporating the paradigm of transformer~\cite{vaswani_attention_2017}. Specifically, an input $\mathbf{X}^{l}=(\mathbf{X}_0^{l}, \mathbf{X}_1^{l}, \cdots, \mathbf{X}_{h-1}^{l})$ with $h$ heads in channel dimension first goes through a multi-head convolution block. The $i$-th input $\mathbf{X}_i^{l}$ is fed into the $i$-th single-head convolutional block, which consists of a vanilla convolutional layer, a BN layer, and a rectified linear unit (ReLU) layer~\cite{nair2010rectified}. Then, the outputs of $h$ heads are concatenated and are further interacted by a pointwise convolutional layer~\cite{hua2018pointwise}. A residual structure with skip connection~\cite{he2016deep} is then utilized to suppress the vanishing gradient problem. Finally, a multilayer perceptron (MLP) layer with another residual structure is applied to obtain local feature.

Our proposed local relational aggregator inherits the advantage of convolution that can aggregate contexts in local regions with efficient computations, in which local token affinity is captured with a small amount of parameters.

\subsubsection{Global Relational Aggregator}

Besides the capture of local details, it is also important to exploit global correlations in the broader token space. The architecture of our proposed global relational aggregator is illustrated in the bottom of Fig.~\ref{fig:overview}. It is composed of a pointwise convolutional layer, a multi-head self-attention block~\cite{vaswani_attention_2017}, a multi-head convolution block, and a MLP layer, in which three residual structures are adopted. 

Denote the input of the multi-head self-attention block be $\mathbf{X}^{g}=(\mathbf{X}_0^{g}, \mathbf{X}_1^{g}, \cdots, \mathbf{X}_{h-1}^{g})$. For the $i$-th head, we first calculate the queries $\mathbf{Q}_i$, keys $\mathbf{K}_i$, and values $\mathbf{V}_i$ as
\begin{subequations}
\label{eq:to_qkv}
\begin{equation}
\mathbf{Q}_i = \mathbf{W}_{\mathbf{Q}_i}\mathbf{X}_i^{g},
\end{equation}
\begin{equation}
\mathbf{K}_i = \mathbf{W}_{\mathbf{K}_i}\mathbf{X}_i^{g},
\end{equation}
\begin{equation}
\mathbf{V}_i = \mathbf{W}_{\mathbf{V}_i}\mathbf{X}_i^{g},
\end{equation}
\end{subequations}
where $\mathbf{W}_{\mathbf{Q}_i}$, $\mathbf{W}_{\mathbf{K}_i}$, and $\mathbf{W}_{\mathbf{V}_i}$ are learnable weight matrices. To map $\mathbf{Q}_i$ and $\mathbf{K}_i$-$\mathbf{V}_i$ pair to a new output, the self-attention is defined as
\begin{equation}
\label{eq:attention}
\mathbf{A}_i=\sigma(\frac{\mathbf{Q}_i{\mathbf{K}_i}^T}{\sqrt{dim}})\mathbf{V}_i ,
\end{equation}
where $\mathbf{Q}_i$, $\mathbf{K}_i$, and $\mathbf{V}_i$ have the same channel dimension $dim$, $\frac{1}{\sqrt{dim}}$ is adopted to scale the dot product, and $\sigma(\cdot)$ is a Softmax function for weighted summing of the values $\mathbf{V}_i$.
Then, a feed forward network is applied to each spatial position for further encoding:
\begin{equation}
\label{eq:ffn}
\mathbf{Y}_i^{g}=FC(\mathbf{A}_i),
\end{equation}
where $FC(\cdot)$ denotes a fully-connected layer, and one dropout layer~\cite{srivastava2014dropout} following the fully-connected layer is omitted. 

The outputs of multiple heads are further fused to be the final output of the multi-head self-attention block. This block and the multi-head convolution block are cooperated to capture global dependencies, and the MLP layer is used to extract the final feature. 

Our global relational aggregator can adaptively model long-range dependencies from distant regions by inheriting the self-attention~\cite{vaswani_attention_2017} paradigm. By progressively stacking local and global relational aggregators, our local-global feature-aware transformer with merits of transformer and convolution can extract complete local-global feature.

\subsection{\Highlight{Rank Scorer and Temporal Pooling}}
\Highlight{A video is ordered sequences of frames, where the frame order also dictates the evolution of the frame appearances~\cite{fernando2016rank}. To model the latent evolution information in frames, we introduce a linear function based rank scorer $\mathbf{u}$. Considering a pair of independent frame representations $\mathbf{F}_i$ and $\mathbf{F}_j$, we aim to learn $\mathbf{u}$ such that $i<j \Leftrightarrow S(\mathbf{F}_i)<S(\mathbf{F}_j)$.
$S(\mathbf{x})$ denotes the rank score, which is defined as}
\begin{equation}
\label{eq:rank_score}
\Highlight{S(\mathbf{x}) = \mathbf{u}^T \cdot \mathbf{x}.}
\end{equation}
\Highlight{Our optimization goal is to make the rank score increase in chronological order. Thus, the rank loss is defined as}
\begin{equation}
\label{eq:rank_loss}
\Highlight{\mathcal{L}_{Rank}=\sum^{t-1}_{k=0}\left |K(k)- S(\mathbf{F}_{k})  \right | ,}
\end{equation}
\Highlight{where $K(\cdot)$ denotes direct proportionality function with a positive parameter. Simultaneously, the frame representations $\{\mathbf{F}_{0}, \mathbf{F}_{1}, \cdots, \mathbf{F}_{t-1}\}$ are reshaped into two-dimensional feature maps and concatenated in chronological order. Then, the temporal pooling achieved through 3D convolution is applied to obtain the dynamic representation $\mathbf{F}^{(d)}$.}

\subsection{Joint Learning of Tasks}


\subsubsection{Dynamic Image Construction}
The dynamic image summarizes the appearances and dynamics of a whole video as one image. The introduction of the dynamic image construction task allows our framework to better extract dynamic features in a ME video, so as to facilitate the performance of MER.

\begin{figure}
\centering\includegraphics[width=0.82\linewidth]{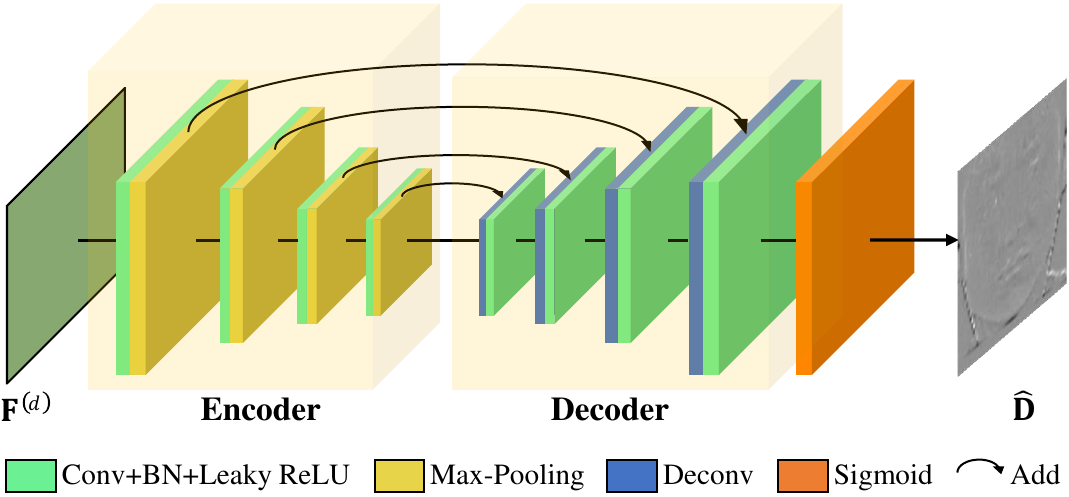} 
\caption{The structure of dynamic image construction module. It is an encoder-decoder as a fully convolutional network without fully-connected layers. The curved arrow denotes skip connection, in which the encoder feature map is element-wise added to the decoder feature map.
}
\label{fig:dynamic}
\end{figure}

The detailed structure of the dynamic image construction module is shown in Fig.~\ref{fig:dynamic}. It contains an encoder network and a decoder network, which is a fully convolutional network without fully-connected layers. This design of full convolution is beneficial for element-wise prediction.
Its end is a Sigmoid layer to produce the estimated single-channel dynamic image $\widehat{\mathbf{D}}$, in which each element in the output of the decoder network is mapped into $(0,1)$ interval. 

The encoder network consists of four consecutive encoder blocks, each of which is composed by a convolutional layer and a max-pooling layer. Particularly, the former is followed by a BN layer and a leaky ReLU layer~\cite{maas2013rectifier}. The decoder network with four decoder blocks is the counterpart of the encoder network. Each decoder block contains a deconvolutional layer and a convolutional layer followed by BN and leaky ReLU. The deconvolutional layer is used to upsample feature maps, which is treated as the inverse process of the max-pooling layer. The skip connections~\cite{he2016deep} between encoder blocks and decoder blocks are beneficial for exploiting encoder information in the decoding process and accelerating the training procedure.



The dynamic image construction loss is defined as
\begin{equation}
\label{eq:dynamic_loss}
\mathcal{L}_{DIC}=MSE(\widehat{\mathbf{D}},\mathbf{D}),
\end{equation}
where $\mathbf{D}$ denotes the ground-truth dynamic image of input video clip $\{\mathbf{I}_{0}, \mathbf{I}_{1}, \cdots, \mathbf{I}_{t-1}\}$, and $MSE(\cdot)$ denotes mean squared error (MSE) loss.

\subsubsection{Micro-Expression Recognition}
The overall architecture of the MER module is illustrated in the right side of Fig.~\ref{fig:overview}. It consists of a max-pooling layer and two fully-connected layers, in which the former is utilized to reduce the dimensions of the dynamic representation $\mathbf{F}^{(d)}$ while maintaining important information, and the latter is used for ME classification. The MER loss is defined as a cross-entropy loss:
\begin{equation}
\label{eq:ME_loss}
\mathcal{L}_{MER}=-\sum^{m-1}_{j=0} p_j\log(\hat{p}_{j}),
\end{equation}
where $m$ denotes the number of ME categories, and $\hat{p}_{j}$ denotes the predicted probability that the video sample is in the $j$-th category. $p_{j}$ denotes the ground-truth probability, which is $1$ if the video sample is in the $j$-th category and is $0$ otherwise.

\subsubsection{Full Loss}
In our joint learning framework, the full loss is combined by $\mathcal{L}_{MER}$,  $\mathcal{L}_{DIC}$ and $\mathcal{L}_{Rank}$ :
\begin{equation}
\label{eq:total_loss}
\Highlight{\mathcal{L}=\mathcal{L}_{MER}+\lambda_{d}\mathcal{L}_{DIC} + \lambda_{r}\mathcal{L}_{Rank},}
\end{equation}
\Highlight{where $\lambda_{d}$ and $\lambda_{r}$ are parameters to weigh the importance of MER, dynamic image construction and rank tasks.}


\begin{table}
\centering\caption{The number of videos for each ME category in CASME II~\cite{yan2014casme}, SAMM~\cite{davison2016samm}, \highlight{and CAS(ME)$^3$~\cite{li2022cas}}. The used five categories \highlight{of CASME II and SAMM, as well as used seven categories of CAS(ME)$^3$} are highlighted with its number in bold. ``-'' denotes this category is not included.}
\label{tab:5-classes distribution}
\setlength\tabcolsep{22pt}
\begin{tabular}{c|*{3}{c}}
\toprule
\diagbox{Class}{Dataset} & CASME II & SAMM &\highlight{CAS(ME)$^3$}\\
\midrule
Happiness &\textbf{32} &\textbf{26} &\textbf{64} \\
Anger &- &\textbf{57}&\textbf{70} \\
Contempt &- &\textbf{12}&- \\
Disgust &\textbf{63}  &9&\textbf{281}\\
Fear &2 &8&\textbf{93} \\
Repression &\textbf{27} &-&-\\
Surprise &\textbf{28}  &\textbf{15}&\textbf{201}\\
Sadness &4 &6&\textbf{64} \\
Others &\textbf{99}  &\textbf{26}&\textbf{170}\\
\bottomrule
\end{tabular}
\end{table}

\begin{table}
\centering\caption{The number of videos for each ME category in CASME II~\cite{yan2014casme},and SAMM~\cite{davison2016samm}, in terms of three-classes evaluation, as well as  CAS(ME)$^2$~\cite{qu2017cas} in terms of four-classes evaluation.}
\label{tab:3-classes distribution}
\setlength\tabcolsep{22pt}
\begin{tabular}{c|*{3}{c}}
\toprule
\diagbox{Class}{Dataset} & CASME II & SAMM &\highlight{CAS(ME)$^2$}\\
\midrule
Positive &32 &26  &8\\
Surprise &28 &15  &9\\
Negative &96 &92  &21\\
Others &- &- &19\\
\bottomrule
\end{tabular}
\end{table}

\section{Experiments}\label{sec4}

\subsection{Datasets and Settings}

\subsubsection{Datasets}

We evaluate our method on \highlight{four} popular spontaneous ME datasets, including CASME II~\cite{yan2014casme}, SAMM~\cite{davison2016samm}, \highlight{CAS(ME)$^2$~\cite{qu2017cas}, and CAS(ME)$^3$~\cite{li2022cas}.}

    $\bullet$ \textbf{CASME II} consists of $255$ videos captured from $26$ subjects in steady and high-intensity illumination. To elicit the MEs, subjects are induced to experience a high arousal with motivations to disguise. Each video is recorded with the frame rate of $200$ frames per second (FPS) and the frame size of $280\times 340$. \hhhighlight{The average duration of MEs is 66.21 frames.} Following the previous methods~\cite{lei2020novel,xia2021micro}, we use ME categories of happiness, disgust, repression, surprise, and others for five-classes evaluation, and use ME categories of positive, negative, and surprise for three-classes evaluation.
    
    $\bullet$ \textbf{SAMM} includes $159$ videos at $200$ FPS from $29$ subjects, which are collected using gray-scale cameras in constrained lighting conditions without flickering. The MEs are elicited from stimuli tailored to each subject. \hhhighlight{The average duration of MEs is 74.31 frames.} Similar to the previous works~\cite{lei2020novel,xia2021micro}, we select ME categories of happiness, anger, contempt, surprise, and others for five-classes evaluation, and select ME categories of positive, negative, and surprise for three-classes evaluation.
    
    $\bullet$ \highlight{\textbf{CAS(ME)$^2$} contains $87$ long videos, each of which includes spontaneous macro-expressions or MEs. These videos are further cropped as $300$ macro-expression video clips and $57$ ME video clips. \hhhighlight{The average duration of MEs is 12.58 frames.} We only evaluate on the ME video clips, in terms of four classes (positive, negative, surprise, and others). 
    }
    
    $\bullet$ \highlight{\textbf{CAS(ME)$^3$} provides $1,109$ MEs and $3,490$ macro-expressions from $100$ subjects, in which each subject is asked to watch $13$ emotionally stimuli and keep their faces expressionless. 
    The recorded videos have the resolution of $1,280 \times 720$. \hhhighlight{The average duration of MEs is 28.61 frames.}
    We conduct experiments on $943$ ME videos from Part A, with seven categories (happiness, disgust, surprise, anger, fear, sadness and others).}

All ME video clips in these datasets are labeled. The number of samples for each ME category are summarized in Table~\ref{tab:5-classes distribution} and Table~\ref{tab:3-classes distribution}, \hlight{and the attributes of each dataset are shown in Table~\ref{tab:attr}.} The dynamic image of each video in these datasets is generated using~\cite{bilen2017action} as the ground-truth annotation.

\begin{table}
\centering\caption{\hlight{The attributes of CASME II~\cite{yan2014casme}, SAMM~\cite{davison2016samm}, CAS(ME)$^2$~\cite{qu2017cas}, and CAS(ME)$^3$~\cite{li2022cas}.}}
\label{tab:attr}
\setlength\tabcolsep{14pt}
\hlight{
\begin{tabular}{c|*{4}{c}}
\toprule
\diagbox{Attribute}{Dataset} & CASME II & SAMM &CAS(ME)$^2$&CAS(ME)$^3$\\
\midrule
Number of Subject &26 &29 &22 &100\\
Frames Per Second &200&200 &30  &30\\
Number of ME Samples &255 &159 &57 &1109\\
Average Duration (frames)&66.21 &74.31 &12.58  &28.61\\
\bottomrule
\end{tabular}
}
\end{table}

\subsubsection{Evaluation Metrics}
Similar to most previous works~\cite{lei2020novel,xia2021micro}, leave-one-subject-out (LOSO) cross-validation is applied in the single dataset evaluation, in which each subject is taken as the test set in turn while the remaining subjects are taken as the training set. We report popular metrics, including accuracy (Acc) and F1-score (F1) \highlight{for CASME II, SAMM, and CAS(ME)$^2$, as well as F1 and unweighted average recall (UAR) for CAS(ME)$^3$}. UAR is defined as
\begin{equation}
    \label{UAR}
    UAR=\frac{1}{m}\sum_{j=0}^{m-1}\frac{TP_j}{TP_j+FN_j},
\end{equation}
where $m$ is the total number of ME categories, 
and $TP_j$, $FP_j$, and $FN_j$ denote the number of true positives, false positives, and false negatives for the $j$-th category, respectively. 

To investigate the generalization ability of our method, we also perform a cross-dataset evaluation. We conduct a two-fold cross-validation on CASME II and SAMM datasets, in which one dataset is used for training while the other dataset is used for testing. Following the settings in previous approaches~\cite{zhao2007dynamic,liu2016MDMO,khor2018enriched}, we report two metrics of weighted average recall (WAR) and UAR. WAR is defined as
\begin{equation}
    \label{WAR}
    WAR=\sum_{j=0}^{m-1}\frac{TP_j}{N},
\end{equation}
where $N$ denotes the total number of samples.

In the following sections, Acc, F1, WAR, and UAR results are all reported in percentages, in which $\%$ is omitted for simplicity.

\subsubsection{Implementation Details}

In our experiments, we extract a video clip with $t$ frames as the input of our FDP by uniformly-space sampling from the raw video.
Each frame image \hhhighlight{is cropped with few background regions and} is aligned to $3 \times 72 \times 72$ via similarity transformation, in which facial shape is preserved without changing the ME.
During training, each image is randomly cropped into $3\times64\times 64$ and is further horizontally flipped to improve the diversity of training data. During testing, each image is centrally cropped into $3\times64\times 64$ so as to be consistent with the training input size.

Our FDP is implemented based on PyTorch~\cite{paszke2019pytorch}, with a solver of Adam~\cite{kingma2014adam}, an initial learning rate of $1\times 10^{-4}$, and a mini-batch size of $36$. The number of frames in the input video clip is set as $t=8$, \hhhighlight{in which each clip is uniformly-spaced sampled from the raw video with a random offset}. \Highlight{The trade-off parameter $\lambda_d$ and $\lambda_r$ are set to $100$ and $0.1$, respectively.} The structure parameters of local-global feature-aware transformer are set as: $N_S=4$, $N_L=2$, and $N_G=1$. All the experiments are conducted on a single NVIDIA GeForce RTX 3090 GPU. FDP takes about $5.8$ GB GPU memory for about $3.5$ hours during training, which demonstrates light-weight transformer structure in our method.

\begin{table}
\centering\caption{Comparison with state-of-the-art methods on CASME II~\cite{yan2014casme} and SAMM~\cite{davison2016samm} for five categories. DL, NDL, PF, RI, and KF denote deep learning based methods, non-deep learning based methods, pre-extracted hand-crafted features, raw images, and key frames, respectively. ``-'' denotes the result is not reported in its paper. The best results are highlighted in bold, and the second best results are highlighted by an underline.}
\label{tab:comparison CAMSE2 and SAMM}
\setlength\tabcolsep{11pt}
\begin{tabular}{*{3}{c}|*{2}{c}|*{2}{c}}

\toprule
\multirow{2}{*}{Method}& \multirow{2}{*}{{Paper}} & \multirow{2}{*}{{Type}} & \multicolumn{2}{c|}{CASME II} & \multicolumn{2}{c}{SAMM}\\
\cmidrule{4-7}
~&~&~ &Acc & \highlight{F1} &Acc &\highlight{F1}\\
\midrule
SparseSampling~\cite{le2017sparsity} &TAFFC'17 & NDL  &49.00 &51.00  &- &-\\

Bi-WOOF~\cite{liong2018less} &SPIC'18 & NDL+KF &58.85 &61.00 &- &- \\

HIGO+Mag~\cite{li2018towards} &TAFFC'18 & NDL  &67.21 &- &- &-\\
FHOFO~\cite{happy2019fuzzy} &TAFFC'19 & NDL &56.64 &52.48 &- &-\\
DSSN~\cite{khor2019dual} &ICIP'19 & DL+PF+KF  &70.78 &72.97  &57.35 &46.44\\

Graph-TCN~\cite{lei2020novel} & MM'20 & DL+RI+KF  &73.98 &72.46  &75.00 &69.85 \\

MicroNet~\cite{xia2020learning} &MM'20 &DL+RI+KF   &75.60 &70.10  &74.10 &73.60 \\
LGCcon~\cite{li2021joint} &TIP'21 &DL+PF &62.14 &60.00 &35.29 &23.00\\
AU-GCN~\cite{lei2021micro} & CVPRW'21 & DL+PF+KF &74.27 &70.47 &74.26 &70.45 \\

GACNN~\cite{kumar2021micro}&CVPRW'21& DL+PF&81.30&70.90 &\textbf{88.24} &\underline{82.79} \\

GEME~\cite{nie2021geme} & NeuCom'21 &DL+PF &75.20 &73.54 &55.88 &45.38 \\

MERSiamC3D~\cite{zhao2021two} &NeuCom'21 &DL+PF+KF &\underline{81.89} &\underline{83.00} &68.75 &64.00 \\

MiNet\&MaNet~\cite{xia2021micro} &IJCAI'21 & DL+RI &79.90 &75.90 &76.70 &76.40 \\

MER-Supcon~\cite{zhi2022micro} &PRL'22 & DL+PF+KF &73.58 &72.86 &67.65 &62.51\\

AMAN~\cite{wei2022novel} &ICASSP'22 &DL+RI &75.40 &71.25& 68.85 &66.82 \\
SLSTT~\cite{zhang2022short} & TAFFC'22 &DL+PF &75.81 &75.30 &72.39 &64.00\\
Dynamic~\cite{sun2022dynamic} &TAFFC'22 &DL+RI+KF  &72.61 &67.00 &-&-\\ 


I$^2$Transformer~\cite{shao2023identity} &APIN'23 &DL+PF+KF &74.26 &77.11  &68.91 &73.01\\

\textbf{FDP} &Ours &DL+RI &\textbf{88.42}&\textbf{87.05} &\underline{86.69}&\textbf{85.29} \\
\bottomrule
\end{tabular}
\end{table}

\subsection{Comparison with State-of-the-Art Methods}
We compare our FDP with state-of-the-art MER methods under the same evaluation setting. These methods can be classified into non-deep learning (NDL) based methods and deep learning (DL) based methods. The latter can be further categorized into pre-extracted feature (PF) based methods and raw image (RI) based methods, according to the type of network input. 

In particular, NDL based methods include LBP-TOP~\cite{zhao2007dynamic}, 3DHOG~\cite{Polikovsky20093dg}, MDMO~\cite{liu2016MDMO}, SparseSampling~\cite{le2017sparsity}, Bi-WOOF~\cite{liong2018less}, HIGO+Mag~\cite{li2018towards}, and FHOFO~\cite{happy2019fuzzy}. DL+PF based methods include \highlight{AlexNet~\cite{krizhevsky2012imagenet}, } Khor \etal~\cite{khor2018enriched}, DSSN~\cite{khor2019dual}, \highlight{STSTNet~\cite{liong2019shallow}, RCN~\cite{xia2020revealing},}  LGCcon~\cite{li2021joint}, AU-GCN~\cite{lei2021micro},
GACNN~\cite{kumar2021micro}, GEME~\cite{nie2021geme}, MERSiamC3D~\cite{zhao2021two}, MER-Supcon~\cite{zhi2022micro}, SLSTT~\cite{zhang2022short}, \highlight{ FR~\cite{zhou2022feature}, HTNet~\cite{wang2023htnet},} and I$^2$Transformer~\cite{shao2023identity}. DL+RI based methods include Peng \etal~\cite{Peng_macro-micro_2018}, Graph-TCN~\cite{lei2020novel}, MicroNet~\cite{xia2020learning},  MiNet\&MaNet~\cite{xia2021micro}, AMAN~\cite{wei2022novel}, and Dynamic~\cite{sun2022dynamic}.
Besides, some of these methods rely on key frames (KF) of MEs, or employ outside training data such as macro-expression datasets.

\subsubsection{Single Dataset Evaluation}
Table~\ref{tab:comparison CAMSE2 and SAMM} and Table~\ref{tab:comparison CAMSE2 and SAMM 3cls} show the comparison results on single datasets of CAMSE II and SAMM for five categories and three categories, respectively. It can be seen that DL based methods often outperform NDL based methods, which proves the power of deep networks.
Note that some recent state-of-the-art methods like GACNN~\cite{kumar2021micro} \highlight{achieve excellent results. This is mainly because these methods rely on auxiliary information such as hand-crafted features and key frames, which assist them to capture ME related information.
}
In contrast, our FDP is significantly better on \highlight{most evaluations} by directly processing raw images. Besides, compared to the methods like MiNet\&MaNet using additional macro-expression datasets, FDP performs better with only benchmark training samples.

\highlight{Moreover, we evaluate our method on more challenging datasets CAS(ME)$^2$ and CAS(ME)$^3$ in Table~\ref{tab:casme^2} and Table~\ref{tab:comparison CAMSE3}, respectively. Note that CAS(ME)$^2$ and CAS(ME)$^3$ datasets exhibit more varieties than CASME II and SAMM datasets.
} 
\hhhighlight{The recent released CAS(ME)$^3$ has the largest number of samples. Compared with CAMSE II and SAMM, it has an abundant number of samples in all seven categories, all of which can be used for training. Meanwhile, CAS(ME)$^3$ is also the most challenging one because its data contains more noise compared to other datasets. 
In this challenging case,} our FDP \hhhighlight{still} significantly outperforms other methods.

\begin{table}
\centering\caption{Comparison with state-of-the-art methods on CASME II~\cite{yan2014casme} and SAMM~\cite{davison2016samm} for three categories. The best results are highlighted in bold. 
}
\label{tab:comparison CAMSE2 and SAMM 3cls}
\setlength\tabcolsep{11.5pt}
\begin{tabular}{*{3}{c}|*{2}{c}|*{2}{c}}

\toprule
\multirow{2}{*}{Method}& \multirow{2}{*}{{Paper}} & \multirow{2}{*}{{Type}} & \multicolumn{2}{c|}{CASME II} & \multicolumn{2}{c}{SAMM}\\
\cmidrule{4-7}
~&~&~ &Acc & \highlight{F1}  &Acc &\highlight{F1}\\
\midrule

OFF-ApexNet~\cite{gan2019off} &SPIC'19  &DL+PF+KF  &88.28 &86.97  &68.18 &54.23 \\

AU-GACN~\cite{xie2020assisted}&MM'20 &DL+RI  &71.20 &35.50  &70.20 &43.30\\ 
 
GACNN~\cite{kumar2021micro} &CVPRW'21 & DL+PF  &89.66 &86.95  &88.72 &81.18 \\

MER-Supcon~\cite{zhi2022micro} &PRL'22 & DL+PF+KF  &89.65 &88.06 &81.20 &71.25\\


\textbf{FDP} &Ours & DF+RI &\textbf{92.72} &\textbf{90.71}  &\textbf{91.25} &\textbf{86.67} \\

\bottomrule
\end{tabular}
\end{table}

\begin{table}
\centering\caption{\highlight{Comparison with state-of-the-art methods on CAS(ME)$^2$~\cite{qu2017cas}. The reported results of LBP-TOP are from~\cite{qu2017cas}, and other methods are implemented using its released code. The best results are highlighted in bold.}}
\label{tab:casme^2}
\setlength\tabcolsep{18pt}
\begin{tabular}{*{3}{c}|*{2}{c}}
\toprule
Method & Paper & Type & Acc & F1\\
\midrule
LBP-TOP~\cite{zhao2007dynamic}&TPAMI'07 &NDL &40.95 &-\\
MicroExpSTCNN~\cite{reddy2019spontaneous}&IJCNN'19&DL+RI &67.35 &54.43\\
AU-GCN~\cite{lei2021micro}& CVPRW'21 & DL+PF+KF &69.38 &65.21\\
SLSTT~\cite{zhang2022short}& TAFFC'22 &DL+PF &75.51 &73.98\\
\textbf{FDP}&Ours&DL+RI &\textbf{83.67} &\textbf{81.69}\\
\bottomrule
\end{tabular}
\end{table}

\begin{table}
\centering\caption{\highlight{Comparison with state-of-the-art methods on CAS(ME)$^3$~\cite{li2022cas}.The results of previous methods except for HTNet~\cite{wang2023htnet} are reported by~\cite{li2022cas}. The best results are highlighted in bold
}
}
\label{tab:comparison CAMSE3}
\setlength\tabcolsep{22.3pt}
\begin{tabular}{*{3}{c}|*{2}{c}}

\toprule
Method& Paper & Type &F1 & UAR\\
\midrule
AlexNet~\cite{krizhevsky2012imagenet} &NeurIPS'12 & DL+KF &25.70 &26.34  \\
STSTNet~\cite{liong2019shallow} &FG'19 & DL+PF+KF &37.95 &37.92  \\
RCN~\cite{xia2020revealing} &TIP'20 & DL+PF+KF &39.28 &38.93 \\
FR~\cite{zhou2022feature} &PR'22 & DL+PF+KF &34.93 &34.13 \\
HTNet~\cite{wang2023htnet} &arXiv'23 &DL+PF+KF &57.67 &54.15\\ 
\textbf{FDP} &Ours &DL+RI &\textbf{59.78}&\textbf{57.84} \\
\bottomrule
\end{tabular}
\end{table}

\begin{table}
\centering
\caption{WAR and UAR results for three ME categories (happiness, surprise, and others) of cross-dataset evaluations. Avg. denotes the average results of two cross-dataset evaluations. The results of methods except for I$^2$Transformer~\cite{shao2023identity} are reported by~\cite{yap2018facial}. CASME II$\rightarrow$SAMM denotes training on CASME II and testing on SAMM. The best results are highlighted in bold, and the second best results are highlighted by an underline.}
\label{tab:compare_cross}
\begin{tabular}{*{3}{c}|*{2}{c}|*{2}{c}|*{2}{c}}
\toprule
\multirow{2}{*}{Method} &\multirow{2}{*}{Paper} & \multirow{2}{*}{{Type}} & \multicolumn{2}{c|}{CASME II$\to$SAMM}& \multicolumn{2}{c|}{SAMM$\to$CASME II}&\multicolumn{2}{c}{Avg.} \\
\cmidrule{4-9}
~&~&~&WAR &UAR &WAR &UAR &WAR &UAR\\
\midrule
LBP-TOP~\cite{zhao2007dynamic}&TPAMI'07 &NDL & 33.8 &32.7 &23.2 &31.6 & 28.5 &32.2\\
3DHOG~\cite{Polikovsky20093dg}&ICDP'09 &NDL & 35.3 &26.9 & 37.3 &18.7 & 36.3 &22.8 \\
MDMO~\cite{liu2016MDMO}&TAFFC'16 &NDL & 44.1 &34.9 & 26.5 &\underline{34.6} & 35.3&34.8\\
Peng \etal~\cite{Peng_macro-micro_2018} &FG'18 &DL+RI+KF & 48.5 &38.2 & 38.4 &32.2 & 43.5 &35.2\\
Khor \etal~\cite{khor2018enriched}&FG'18 &DL+PF+KF & \underline{54.4} &\underline{44.0} & 57.8 &33.7 & 56.1 &\underline{38.9}\\
I$^2$Transformer~\cite{shao2023identity} &APIN'23 &DL+PF+KF & 51.2 &- & \textbf{66.2}&- & \underline{58.7}&-\\
\textbf{FDP}&Ours &DL+RI& \textbf{58.2} &\textbf{51.8} & \underline{62.2} &\textbf{56.0}& \textbf{60.2}&\textbf{53.9}\\
\bottomrule

\end{tabular}
\end{table}


\hlight{It can be observed that our FDP achieves the overall best performance across datasets with varying categories, scales, and noise levels. Specifically, FDP processes raw video sequences, requiring no category-specific adjustments or auxiliary data. This confirms its applicability to any ME-containing video. In addition, by directly processing raw images without dependencies on key frames, pre-extracted features, or external datasets, FDP eliminates pre-processing dependencies. This enables deployment in real-world scenarios where prior information is inaccessible. Therefore, our FDP is a practical MER solution.
}

\subsubsection{Cross-Dataset Evaluation}

Table~\ref{tab:compare_cross} presents the cross-dataset evaluation results. The common three ME categories of happiness, surprise, and others for the two datasets are used. It can be observed that our approach achieves the best average performance especially for the UAR metric,
which demonstrates the strong generalization ability of our FDP. This can be attributed to two merits of our method. First, our proposed local-global feature-aware transformer has strong capacities of relational reasoning and feature learning by simultaneously modeling local and global contexts. Second, the joint learning with dynamic image construction is beneficial for extracting ME related features, and thus improves the robustness on unseen samples. 

\subsubsection{\hlight{Systematic Discussion and Structured Gap Analysis}}

\hlight{
The above results demonstrate that our FDP outperforms state-of-the-art MER methods in terms of both single dataset evaluation and cross-dataset evaluation. There are two main limitations in previous MER methods:
}

\hlight{$\bullet$ \textbf{Dependency on Auxiliary Inputs and Pre-Processing}}
    
    \hlight{\textit{Pre-extracted Feature (PF) Reliance}: Top-performing methods like GACNN~\cite{kumar2021micro}, STSTNet~\cite{liong2019shallow}, RCN~\cite{xia2020revealing}, and FR~\cite{zhou2022feature} rely on pre-extracted optical flow or other hand-crafted features. This introduces complexity and sensitivity to the quality of feature extraction techniques. The need for separate feature computation hinders end-to-end learning and real-time applicability.}

    \hlight{\textit{Key Frame (KF) Reliance}: Previous methods such as DSSN~\cite{khor2019dual}, AU-GCN~\cite{lei2021micro}, MERSiamC3D~\cite{zhao2021two}, MER-Supcon~\cite{zhi2022micro}, and RCN~\cite{xia2020revealing} require accurate detection of key frames. This dependency is problematic when key frames are not provided or not detected correctly in real scenarios, leading to cumulative errors and limited robustness.}

    \hlight{\textit{External Data Dependency}: Existing approaches like MiNet\&MaNet~\cite{xia2021micro} utilize additional macro-expression datasets for training. This reduces practicality, as such data may be not readily available or directly relevant.}

\hlight{$\bullet$ \textbf{Limited Generalization and Robustness} }
    
    \hlight{\textit{Dataset Sensitivity}: The performances of some methods vary across datasets and category amounts. For instance, GACNN~\cite{kumar2021micro} excels on SAMM with five categories but drops significantly on CASME II with five categories. This indicates overfitting to specific dataset characteristics or evaluation benchmarks.}

    \hlight{\textit{Category Scalability}: Existing methods often struggle when the number of ME categories increases. Their performances generally degrade in five-classes evaluation (see Table~\ref{tab:comparison CAMSE2 and SAMM}) compared to three-classes evaluation (see Table~\ref{tab:comparison CAMSE2 and SAMM 3cls}), showing the limitations in feature discriminability and model capacity for recognizing fine-grained categories.} 

    \hlight{\textit{Noise Vulnerability}: The performances often degrade on challenging datasets like CAS(ME)$^3$, where noises are prevalent. Some methods relying on key frames or hand-crafted features are particularly susceptible, demonstrating poor noise robustness.}

\hlight{Existing methods exhibit critical gaps in practicality, robustness, and generalization. Their performances usually rely on pre-processing, external data, or specific dataset conditions. In contrast, our FDP overcomes these limitations, and the results from Tables~\ref{tab:comparison CAMSE2 and SAMM} to \ref{tab:compare_cross} demonstrate the superiority.}

\begin{table}
\centering\caption{Acc and F1 results for different variants of FDP on SAMM~\cite{davison2016samm} in terms of five categories. L: local relational aggregator; G: global relational aggregator. The best results are highlighted in bold.}
\label{tab:ablation study}
\setlength\tabcolsep{49pt}
\begin{tabular}{c|*{2}{c}}
\toprule
Method & Acc & \highlight{F1}\\
\midrule
\textbf{FDP} &\textbf{86.69} &\textbf{85.29}\\
FDP w/o $\mathcal{L}_{DIC}$ &83.98 &80.40\\
\Highlight{FDP w/o $\mathcal{L}_{Rank}$} &83.54 &80.02\\
FDP w/o L &81.79 &79.78\\
FDP w/o G &78.54 &75.32\\
FDP w/o L\&G &60.70 &59.05\\
\bottomrule
\end{tabular}
\end{table}

\subsection{Ablation Study}
\label{ssec:ablation}
In this section, we conduct ablation experiments to investigate the effectiveness of dynamic image construction module, \Highlight{rank scorer,} local-global feature-aware transformer, and backbone structure on MER. The results of different variants of FDP are shown in Table~\ref{tab:ablation study}. 
These experiments are all evaluated on SAMM dataset in terms of five categories.

\subsubsection{Dynamic Image Construction}
Compared with the FDP, the performance of FDP w/o $\mathcal{L}_{DIC}$ is degraded after removing the dynamic image construction module. This demonstrates that dynamic image construction task in our joint learning framework contributes to MER. The estimation of dynamic image can guide the dynamic representation $\mathbf{F}^{(d)}$ shared by the MER module to capture spatial appearances and temporal patterns.

\Highlight{\subsubsection{Rank Scorer}
When removing $\mathcal{L}_{Rank}$ of the FDP, the Acc and F1 results of FDP w/o $\mathcal{L}_{Rank}$ are decreased to $83.54$ and $80.02$, respectively, which shows the effectiveness of rank scorer. This is mainly because the supervision of rank scorer can enhance the model’s understanding on the evolution of micro-expression actions.}

\subsubsection{Local-Global Feature-Aware Transformer}
Here we evaluate the main components of local-global feature-aware transformer, including local relational aggregator and global relational aggregator.
When removing both two relational aggregators, the Acc and F1 results of FDP w/o L\&G are significantly decreased to $60.70$ and $60.15$, respectively. If we remain either relational aggregator, the results improve a lot. However, the performance is still worse than FDP. This demonstrates the effectiveness of local-global feature-aware transformer with local-global relational reasoning and feature learning, which largely determines the performance of FDP as the backbone.

\begin{table}
\centering\caption{\highlight{SAMM~\cite{davison2016samm} results (five categories) and the number of parameters (\#Params.) for different backbone structures of FDP. The best results are highlighted in bold.}}
\label{tab:backbone}
\setlength\tabcolsep{20pt}
\begin{tabular}{c|c|*{3}{c}}
\toprule
Backbone &Type &Acc & F1 &\#Params.\\
\midrule
ViT-Base~\cite{dosovitskiy2020image} &T &70.58 &67.24 &88.39M\\
Ours & C+T &79.82 &74.46 &28.71M \\ 
Ours & C$'$+T &83.93 &79.21 &27.14M \\ 
\textbf{Ours} &C$^*$+T &\textbf{86.69} &\textbf{85.29} &14.82M\\
ResNet18~\cite{he2016deep} &C &76.47 &73.25 &\textbf{13.27M}\\
\midrule

\multicolumn{5}{l}{T: Transformer}\\
\multicolumn{5}{l}{C: Conv.
}\\
\multicolumn{5}{l}{C$'$: Conv. + BN + ReLU + Pointwise Conv.}\\
\multicolumn{5}{l}{C$^*$: Multi-head Conv.}\\
\bottomrule
\end{tabular}
\end{table}

\highlight{\subsubsection{Backbone Structure}
Table~\ref{tab:backbone} shows the number of parameters of different backbone structures, in which the results are obtained by replacing our proposed local-global feature-aware transformer with new backbone. When directly using the classical vision transformer ViT-Base as the backbone, we obtain low Acc and F1 results with a large number of parameters. If using a single vanilla convolutional layer to replace our proposed multi-head convolution block, the performance is significantly improved over ViT-Base. This is because convolution works better than ViT-Base in extracting local features, which demonstrates the importance of local features for MER. When further adding pointwise convolution, the results are better while the number of parameters is slightly decreased. This is attributed to the enhanced feature learning ability and the reduced number of output channels by pointwise convolution.}

\highlight{When changing the single head to multiple heads, our final version of FDP achieves the best performance using the least parameters. This is because our proposed multi-head convolution captures more diverse ME information from the input data by allowing different groups of channels to learn independent features. We also compare with a classical convolutional network ResNet18. Although it requires less parameters, its performance is significantly worse. Compared to typical vision transformers and convolutional networks, our FDP performs better by integrating their both advantages. 
}

\hlight{We also notice that FDP shows markedly superior training efficiency compared to ViT-Base~\cite{dosovitskiy2020image} due to its lightweight hybrid architecture. The multi-head convolution reduces optimization complexity by autonomously capturing diverse spatial representations without manual feature engineering. Besides, the self-stabilizing properties of the pointwise convolution layers reduce the requirement of learning rate scheduling, which substantially diminishes tuning effort. Moreover, FDP exhibits robustness to hyperparameter changing comparing to other variants, demonstrating better hyperparameter insensitivity.}

\hhhighlight{
\subsubsection{Weights of Losses} As shown in Eq.~\eqref{eq:total_loss}, the full loss is composed of MER loss $\mathcal{L}_{MER}$, dynamic image construction loss $\mathcal{L}_{DIC}$, and rank loss  $\mathcal{L}_{Rank}$. 
Table~\ref{tab:weights} shows the results of our FDP using different weights of loss terms.
The first three rows show that when keeping $\lambda_r$ unchanged and $\lambda_d$ increasing, both Acc and F1 results increase.
This is because the numerical scale of $\lambda_d \mathcal{L}_{DIC}$ gradually approaches to that of $\mathcal{L}_{MER}$, enabling each loss term to contribute to the optimization process.
However, when $\lambda_d$ increases to 1000, both Acc and F1 results drop significantly.
Due to the overly large dynamic image construction loss term, $\mathcal{L}_{MER}$ becomes insignificant in the full loss.
When fixing $\lambda_d$ as 100 and changing $\lambda_r$, the similar phenomenon can be found in the last three rows.
Therefore, our method performs the best when the magnitude values of loss terms are balanced.}

\hhhighlight{\subsection{Significance Test}}

\hhhighlight{\subsubsection{Statistical Significance between State-of-the-Art Methods and Our Method}
To prove the significant superiority of our method to previous methods, we conduct a significance test based on the results from Tables~\ref{tab:comparison CAMSE2 and SAMM} to \ref{tab:comparison CAMSE3}. Considering the results of different methods do not follow a normal distribution, we conduct the Wilcoxon rank-sum test~\cite{mann1947test}. Specifically, we make the following hypotheses:
}

    $\bullet$ $H_0$: There is no significant difference between our method and state-of-the-art methods. 
    
    $\bullet$ $H_1$: Our method is significantly superior to state-of-the-art methods. 

\hhhighlight{The test statistics and P-values of the F1 results are shown in Table~\ref{tab:P-value 3-6}. 
It can be seen that all P-values are less than 0.05. Therefore, we reject the null hypothesis $H_0$ and accept the alternative hypothesis $H_1$. Our FDP significantly outperforms state-of-the-art methods in terms of statistics.
}

\begin{table}
\centering\caption{\hhhighlight{
Acc and F1 results for our FDP with different loss term weights on SAMM~\cite{davison2016samm} in terms of five categories. The best results are highlighted in bold.}}
\label{tab:weights}
\setlength\tabcolsep{36.1pt}
\begin{tabular}{*{2}{c}|*{2}{c}}
\toprule
$\lambda_d$ & $\lambda_r$ & Acc & F1\\
\midrule
1 &0.1 &83.98 &80.40\\
10 &0.1 &84.50 &81.19\\
\textbf{100} &\textbf{0.1} &\textbf{86.69} &\textbf{85.29}\\
1000 &0.1 &50.73 &23.67\\
100 &0.01 &83.20 &79.73\\
100 &1 &77.43 &73.89\\
100 &10 &51.47 &35.17\\
\bottomrule
\end{tabular}
\end{table}

\begin{table}
\centering\caption{\hhhighlight{Statistics of Wilcoxon rank-sum test~\cite{mann1947test} and P-values on benchmark F1 results from Tables~\ref{tab:comparison CAMSE2 and SAMM} to~\ref{tab:comparison CAMSE3}.}}
\label{tab:P-value 3-6}
\setlength\tabcolsep{32pt}
\begin{tabular}{c|*{2}{c}}
\toprule
Benchmark  & Statistics & P-value\\
\midrule
CASME II (Five Categories)& 4.977 &3.227e-7\\
SAMM (Five Categories)& 4.243  & 1.106e-5\\
CASME II (Three Categories)& 2.309 & 1.046e-2\\
SAMM (Three Categories)& 2.309 &1.046e-2\\
CAS(ME)$^2$& 1.964 & 2.477e-2\\
CAS(ME)$^3$& 2.611 & 4.512e-3\\
\bottomrule
\end{tabular}
\end{table}

\hhhighlight{
\subsubsection{Statistical Significance between Backbones and Our Proposed Modules}
To investigate the effectiveness of the proposed modules in our framework from the perspective of statistics, we conduct a significance test based on the results from Tables~\ref{tab:ablation study} to \ref{tab:backbone}.
We make the following hypotheses:
}

    $\bullet$ $H_0$: Our framework has no significant effect.
    
    $\bullet$ $H_1$: Our framework has a significant optimization effect.

\hhhighlight{The test statistics and P-values of the F1 results are presented in Table~\ref{tab:P-value 8-9}. 
As can be seen, all P-values are less than 0.05. Therefore, we reject the null hypothesis $H_0$ and accept the alternative hypothesis $H_1$. It is demonstrated that our proposed modules are beneficial for MER.}

\begin{table}
\centering\caption{\hhhighlight{Statistics of Wilcoxon rank-sum test~\cite{mann1947test} and P-values on ablation F1 results from Tables~\ref{tab:ablation study} to~\ref{tab:backbone}.}}
\label{tab:P-value 8-9}
\setlength\tabcolsep{43.5pt}
\begin{tabular}{c|*{2}{c}}
\toprule
Ablation & Statistics & P-value\\
\midrule
Modules & 2.611 &4.512e-3\\
Backbones & 2.309  & 1.046e-2\\
\bottomrule
\end{tabular}
\end{table}

\begin{table}
\centering\caption{Dynamic image construction results (lower is better) for different variants of FDP on SAMM~\cite{davison2016samm}. The best results are highlighted in bold.}
\label{tab:JMD for Dynamic}
\setlength\tabcolsep{57pt}
\begin{tabular}{c|c}
\toprule
Method & Average MSE ($\times 10^{-3}$) \\
\midrule
\textbf{FDP} &\textbf{1.44} \\
FDP w/o $\mathcal{L}_{MER}$ & 1.78\\
FDP w/o L & 6.77\\
FDP w/o G & 6.90\\
FDP w/o L\&G & 7.43\\
\bottomrule
\end{tabular}
\end{table}

\subsection{FDP for Dynamic Image Construction}
We have validated the contribution of dynamic image construction task to MER in Sec.~\ref{ssec:ablation}. To also investigate the effectiveness of MER task for dynamic image construction, we implement a new baseline FDP w/o $\mathcal{L}_{MER}$. It only achieves dynamic image construction by removing the MER module. Besides, FDP w/o L, FDP w/o G, and FDP w/o L\&G are still evaluated to explore the influence of local-global feature-aware transformer on dynamic image construction. We report Average MSE as the evaluation metric, which is computed as the average of MSE between $\mathbf{D}$ and $\widehat{\mathbf{D}}$ over all samples.

Table~\ref{tab:JMD for Dynamic} shows the average MSE on the SAMM benchmark. We can observe that FDP outperforms FDP w/o $\mathcal{L}_{MER}$ with the help of MER. This is attributed to the guidance of the MER task to capture facial subtle muscle actions, which is closely related to the dynamic image. Combining with the observations in Sec.~\ref{ssec:ablation}, it can be concluded that MER and dynamic image construction facilitate each other in our joint learning framework.

Besides, compared with FDP w/o L, FDP w/o G, and FDP w/o L\&G, FDP exhibits a large margin. This demonstrates that our local-global feature-aware transformer is a strong backbone network for capturing spatio-temporal clues.

\begin{figure}
\centering\includegraphics[width=\linewidth]{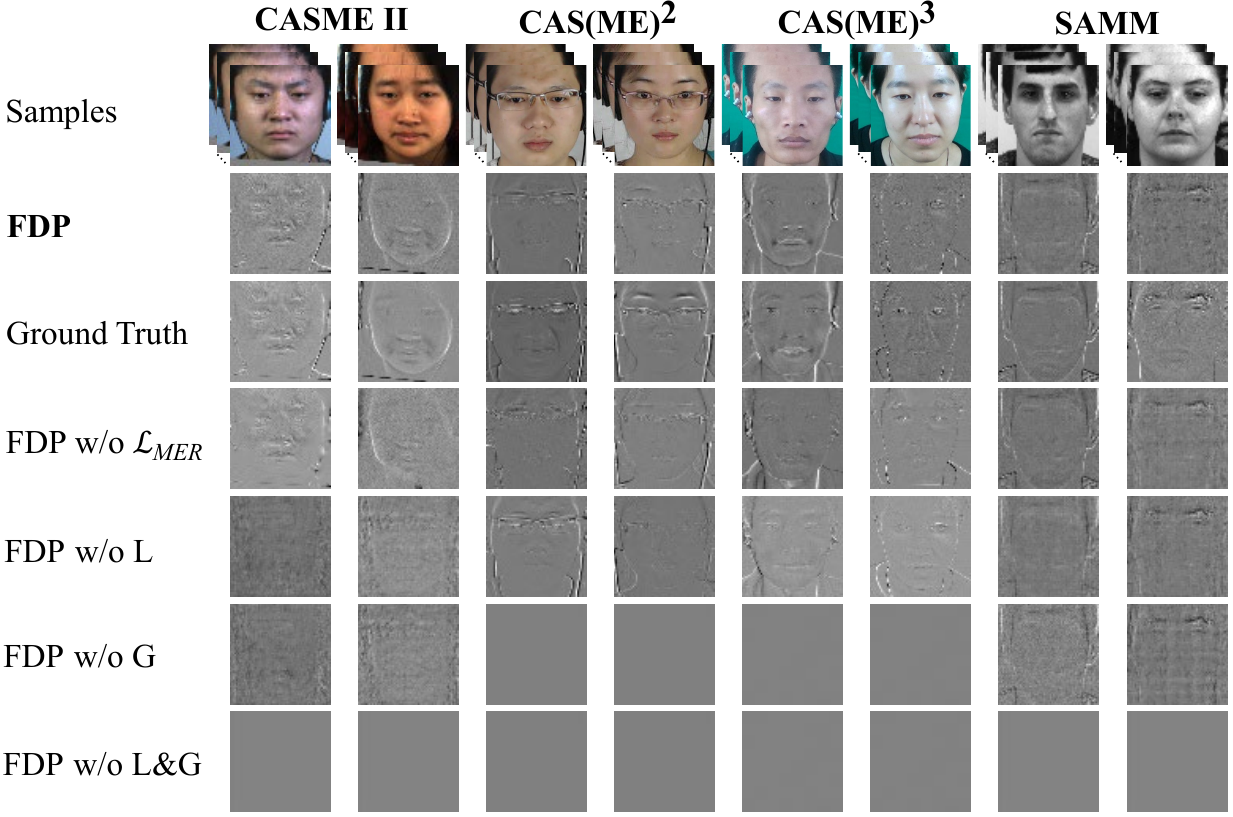} 
\caption{Visualization of dynamic image construction results for example video clips from CASME II~\cite{yan2014casme}, \highlight{CAS(ME)$^2$~\cite{qu2017cas}, CAS(ME)$^3$~\cite{li2022cas}}, SAMM~\cite{davison2016samm}. The third row shows the ground-truth dynamic images, and other rows show the estimated dynamic images of different methods.}
\label{fig:dynamic_viz}
\end{figure}

\subsection{Visual Results}

Fig.~\ref{fig:dynamic_viz} visualizes the dynamic image construction results of these methods on several video clip samples from CASME II, \highlight{CAS(ME)$^2$, CAS(ME)$^3$}, and SAMM. We can see that our FDP extracts dynamic information from ME videos with the best effects. Its estimations are close to the ground-truth annotations, which demonstrates that the fully convolutional encoder-decoder structure of dynamic image construction module is effective for element-wise prediction. Besides, compared to FDP w/o $\mathcal{L}_{MER}$, FDP captures more appearance and motion details. For example, FDP accurately captures the dynamics around eyes for the second video sample, while FDP w/o $\mathcal{L}_{MER}$ fails.

Moreover, the results of FDP and FDP w/o $\mathcal{L}_{MER}$ look more reasonable than FDP variants with incomplete local-global feature-aware transformer. This again proves the effectiveness of local-global feature-aware transformer for the dynamic image construction task. Due to the appearance and motion details captured by the dynamic image construction task, FDP can focus on facial subtle muscle actions associated with MEs. 

\subsection{Limitations}

According to the above experiments, our method significantly outperforms the previous works. 
However, there are a few failure cases, as illustrated in Table~\ref{tab:failure_case}. We notice that mistakenly recognized videos are very challenging, and even their ground-truth dynamic images fail to reflect clear facial subtle motions. For example, the correctly recognized video ``006\_2\_4'' of the subject ``006'' from SAMM has highlighted motions around eyebrows in its dynamic image, while the mistakenly predicted video ``006\_5\_11'' has no significant motions in its dynamic image. 
We will try to solve this challenging motion capture issue in the future work.

\begin{table}
\centering\caption{Failure cases of our FDP on CASME II~\cite{yan2014casme} and SAMM~\cite{davison2016samm}. The incorrect predictions are highlighted in bold. ``DI'' denotes dynamic image.}
\label{tab:failure_case}
\begin{tabular}{c|c|c|c|c|c}
\toprule

\multirow{2}{*}{Subject} & \multicolumn{2}{c|}{Video}& \multicolumn{2}{c|}{Ground Truth} &\multirow{2}{*}{Prediction} \\\cmidrule{2-5}
&Name &Illustration&DI&ME Cate. \\
\midrule
\multirow{2}{*}[-4.5ex]{SAMM~006}&006\_2\_4&
		\raisebox{-.5\height}{\includegraphics[width=0.3\linewidth]{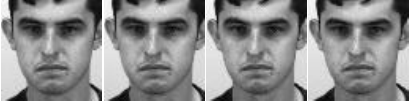}}
	&
		\raisebox{-.5\height}{\includegraphics[width=0.074\linewidth]{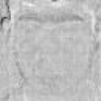}}
	&Anger&Anger\\
~&006\_5\_11&
		\raisebox{-.5\height}{\includegraphics[width=0.3\linewidth]{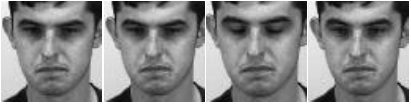}}
	&\raisebox{-.5\height}{\includegraphics[width=0.074\linewidth]{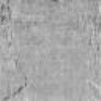}}&Anger&\textbf{Contempt}\\

\midrule
\multirow{2}{*}[-4.5ex]{CASME II~17}&17\_EP03\_09&
		\raisebox{-.5\height}{\includegraphics[width=0.3\linewidth]{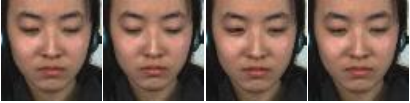}}
	&\raisebox{-.5\height}{\includegraphics[width=0.074\linewidth]{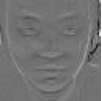}}&Happiness&Happiness\\
~&17\_EP13\_09&
		\raisebox{-.5\height}{\includegraphics[width=0.3\linewidth]{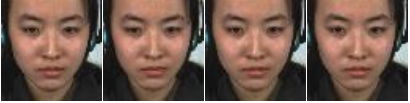}}
	&\raisebox{-.5\height}{\includegraphics[width=0.074\linewidth]{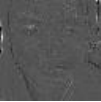}}&Happiness& \textbf{Surprise}\\
\bottomrule
\end{tabular}
\end{table}

\hlight{Besides, our method has limitations when dealing with real-world data. Since most of the existing ME datasets consist of frontal and well-lit face images collected in constrained environments, our method ignores the case of in-the-wild faces. We will also explore more robust techniques to the unconstrained scenarios.}

\section{Conclusion}

\Highlight{In this paper, we have proposed a novel end-to-end fine-grained dynamic perception framework for joint MER and dynamic image construction, in which the rank technique benefits MER and two correlated tasks contribute to each other. Besides, we have developed a local-global feature-aware transformer to extract local-global features. Our framework does not rely on pre-extracted hand-crafted features and key frames, which is a promising solution to MER with good applicability.}

We have compared our method with state-of-the-art works on the challenging CASME II, SAMM, \highlight{CAS(ME)$^2$, and CAS(ME)$^3$} benchmarks. It is shown that our method outperforms previous works for both single dataset evaluation and cross-dataset evaluation. Besides, we have conducted an ablation study which indicates that main components in our framework are all beneficial for MER. Moreover, the experiments on dynamic image construction show excellent performance of our method, and the visual results demonstrate that our method can capture facial subtle muscle actions related to MEs. 

\hhhighlight{In the future work, there are two aspects worthy of further exploring. First, regarding to the process of dynamic image construction, we hope to guide the network to pay more attention to the areas where MEs occur and ignore irrelevant information such as facial shape. 
Therefore, it is promising to develop the technique of disentangle irrelevant information like facial identity information.
Second, the input of our network is a sequence of frames, in which some frames except for the key frames also play an important role in MER. It is promising to design a technique to locate important frames besides the key frames in a video clip, thereby facilitating the MER.
}

\begin{acks}
This work was supported by the National Natural Science Foundation of China (No. 62472424), the China Postdoctoral Science Foundation (No. 2023M732223), the Hong Kong Scholars Program (No. XJ2023037/HKSP23EG01), and the Research Impact Fund of the Hong Kong Government (No. R6003-21). It was also partially supported by the National Natural Science Foundation of China (Nos. 62402231, 72192821, and 62472282), the Opening Fund of the State Key Laboratory of Virtual Reality Technology and Systems, Beihang University (No. VRLAB2024C03), and the Science and Technology Planning Project of Henan Province (No. 242102211003).
\end{acks}

\bibliographystyle{ACM-Reference-Format}
\bibliography{reference}

\appendix

\end{document}